\documentclass[conference]{IEEEtran}
\IEEEoverridecommandlockouts
\usepackage{cite}
\usepackage{amsmath,amssymb,amsfonts}
\usepackage{algorithmic}
\usepackage{graphicx}
\usepackage{textcomp}
\usepackage{xcolor}

\usepackage{multirow}
\usepackage{multicol}
\usepackage{booktabs}
\usepackage{hyperref}
\def\BibTeX{{\rm B\kern-.05em{\sc i\kern-.025em b}\kern-.08em
    T\kern-.1667em\lower.7ex\hbox{E}\kern-.125emX}}
\begin{document}

\title{Knowledge Distillation with Deep Supervision\thanks{This work is supported by National Key R\&D Program of China (Grant No: 2019YFB1600700), the Starry Night Science Fund of Zhejiang University Shanghai Institute for Advanced Study (Grant No: SN-ZJU-SIAS-001), National Natural Science Foundation of China (Grant No: U1866602) and the advanced computing resources provided by the Supercomputing Center of Hangzhou City University.}
}

\author{\IEEEauthorblockN{1\textsuperscript{st} Shiya Luo}
\IEEEauthorblockA{\textit{Zhejiang University} \\
Hangzhou, China \\
lsya@zju.edu.cn}
\and
\IEEEauthorblockN{2\textsuperscript{nd} Defang Chen}
\IEEEauthorblockA{\textit{Zhejiang University} \\
Hangzhou, China \\
defchern@zju.edu.cn}
\and
\IEEEauthorblockN{3\textsuperscript{rd} Can Wang$^{\star}$\thanks{$^{\star}$Corresponding author}\thanks{Can Wang is also with the ZJU-Bangsun Joint Research Center, and Shanghai Institute for Advanced Study of Zhejiang University.}}
\IEEEauthorblockA{\textit{Zhejiang University} \\
\textit{Hangzhou City University}\\
Hangzhou, China \\
wcan@zju.edu.cn}}

\maketitle

\begin{abstract}
Knowledge distillation aims to enhance the performance of a lightweight student model by exploiting the knowledge from a pre-trained cumbersome teacher model. 
However, in the traditional knowledge distillation, teacher predictions are only used to provide the supervisory signal for the last layer of the student model, which may result in those shallow student layers lacking accurate training guidance in the layer-by-layer back propagation and thus hinders effective knowledge transfer. To address this issue, we propose Deeply-Supervised Knowledge Distillation (DSKD), which fully utilizes class predictions and feature maps of the teacher model to supervise the training of shallow student layers. A loss-based weight allocation strategy is developed in DSKD to adaptively balance the learning process of each shallow layer, so as to further improve the student performance. Extensive experiments on CIFAR-100 and TinyImageNet with various teacher-student models show significantly performance, confirming the effectiveness of our proposed method. Code is available at: \href{https://github.com/luoshiya/DSKD}{https://github.com/luoshiya/DSKD}
\end{abstract}

% \begin{IEEEkeywords}
% component, formatting, style, styling, insert
% \end{IEEEkeywords}

\section{Introduction}

Deep neural networks have shown excellent performance in the computer vision tasks with massively parameterized models and huge calculations \cite{Krizhevsky2012ImageNetCW,simonyan2015Very,He2016Deep,Zagoruyko2016Wide}. These expensive computation and storage cost in turn make them difficult to be deployed on mobile devices with limited resources and real-time applications demanding quick response. The recently proposed knowledge distillation (KD) technique provides a possible solution to this problem by training a small student model to mimic the performance of a large teacher model \cite{Hinton2015Distilling,Gou2021KnowledgeDA}. 

In the vanilla knowledge distillation, 
class predictions of the teacher model are exploited to provide the training guidance for the last layer of the student model \cite{Hinton2015Distilling}. 
However, this supervisory signal starting from only the last student layer would gradually weaken as the gradient is back propagated layer-by-layer, leading to the accumulation of training bias in shallow student layers and hurt the effectiveness of knowledge transfer \cite{Lee2015Depply,Sun2019Deeply}. 

To tackle this problem, in this paper, we propose Deeply-Supervised Knowledge Distillation (DSKD) to improve the participation of shallow layers in teacher knowledge transfer. 
Generally, giving shallow layers extra supervisory signals and increasing their discriminative ability can effectively prevent training bias from propagating from the last layer to shallow layers and thus reduce the final prediction error \cite{Lee2015Depply,Szegedy2015Going}.
Such intermediate targets empirically help the model generalize well, which is analogous to human learning that high-level knowledge could be better captured with the help of useful intermediate concepts \cite{gulccehre2016knowledge}. 
Actually, the knowledge learned by shallow layers in our method is rather similar to an intermediate learning clue for the student model training.

% Explain from forward propagation, this is analogous to the human learning process, where learning a certain knowledge directly in the absence of relevant pre-knowledge can lead to insufficient understanding resulting in only a small part being learned. Therefore, pre-knowledge is important for learning the target knowledge, and the knowledge learned in shallow layers is equivalent to the pre-knowledge, so it is more helpful to improve the final performance. 

Since hierarchical concepts contained in intermediate feature maps are beneficial for knowledge transfer \cite{Romero2015FitNets,Zagoruyko2017Paying,Ahn2019Variational,Chen2021Cross}, we also leverage the teacher feature maps, besides class predictions, as another knowledge source to supervise the student training from the last layer to shallow layers.
Additionally, we develop a loss-based weighting strategy to adaptively balance the different learning speeds of those shallow layers. 

To conclude, all contributions of our work are summarized as follows:
\begin{itemize}
    \item We propose a novel technique to further improve the final performance by employing multiple auxiliary classifiers attached to various shallow layers of the student to learn the teacher knowledge.
    \item We introduce a loss-based weighting strategy that adaptively assigns different learning weights to different auxiliary classifiers during training to balance the learning speed of each shallow layer.
    \item The effectiveness of our proposed method is verified in extensive experiments including \textit{eleven} competitors and \textit{seven} groups of the teacher-student architectures.
    
\end{itemize}

\begin{figure*}[t]
  \centering
  \includegraphics[width=0.95\textwidth]{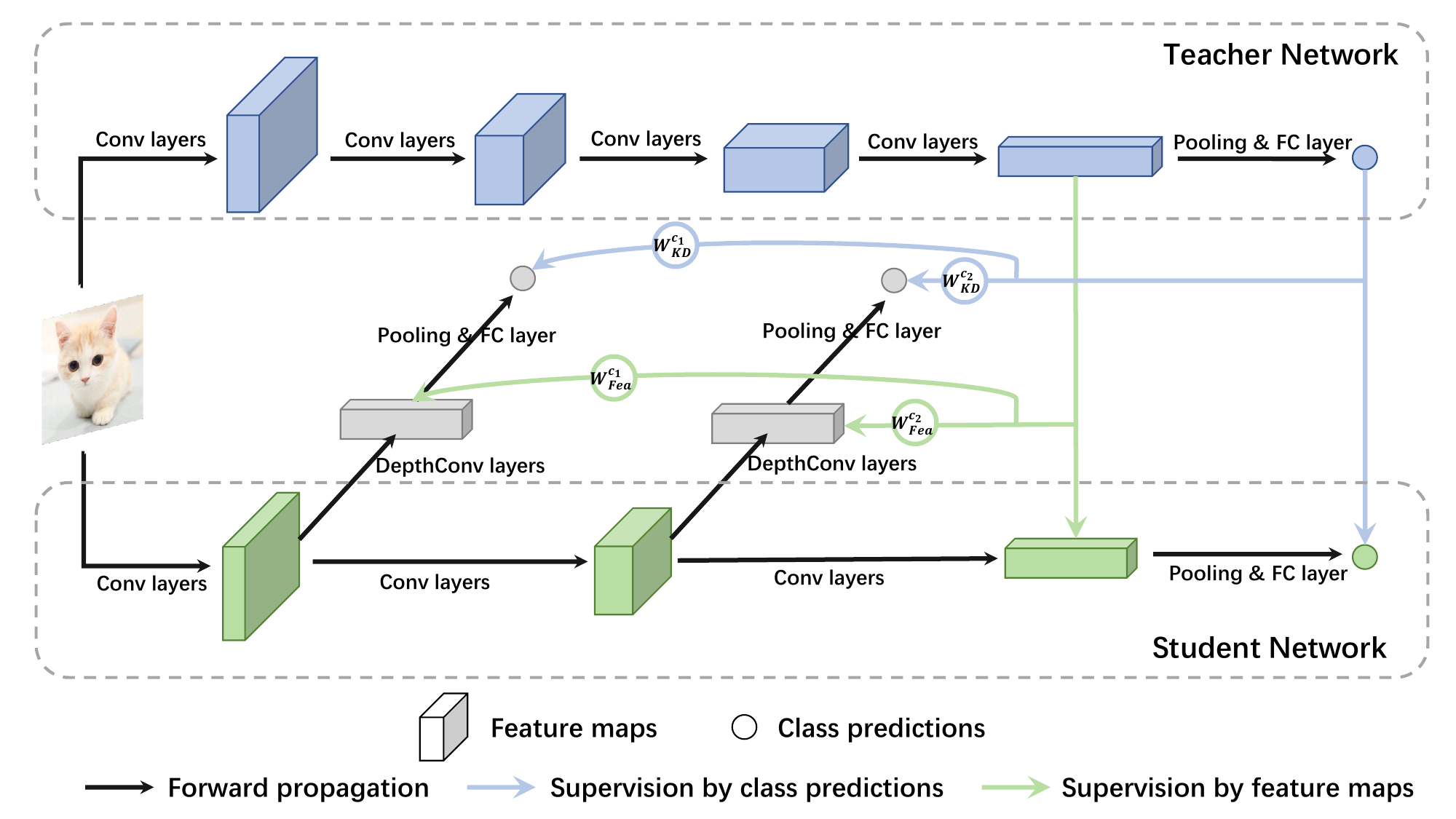}
  \caption{An overview of our proposed Deeply-Supervised Knowledge Distillation (DSKD). We use a toy example containing only two shallow layers for illustration. Each shallow student layer is equipped with an auxiliary classifier for learning class predictions and feature maps of the pre-trained teacher model. Adaptive weights are assigned to different auxiliary classifiers to balance the learning process.
  }
  \label{fig:overview}
% \vspace{-0.3cm}
\end{figure*}

\section{Related Work}
{\bf Knowledge distillation.} 
Knowledge distillation (KD) is proposed to distill  knowledge from a large teacher model into a small student model, serving as a model compression technique \cite{Hinton2015Distilling,Ji2021Show, chen2020online,chen2022knowledge}. The vanilla KD  \cite{Hinton2015Distilling} employs class predictions from the last fully-connected layer of the teacher as the teacher knowledge to force that of the student to match it.  
Besides class predictions, many subsequent works are devoted to exploring different useful teacher knowledge, such as features and relations. 

Feature distillation excavates the knowledge in the output of intermediate layers to supervise the training of the student \cite{Romero2015FitNets,Zagoruyko2017Paying,Ahn2019Variational,Tian2020ContrastiveRD,Liu2021ExploringIC,Chen2021Cross,lin2022knowledge}. For example, FitNet \cite{Romero2015FitNets}  proposes to simply transfer feature maps between a selected teacher-student layer pair for better final performance.  AT \cite{Zagoruyko2017Paying} develops attention maps derived from feature maps to transfer knowledge between all layer pairs in a one-to-one manner.  Relation distillation constructs the relationship between different samples or layers for knowledge transfer \cite{Tung2019SimilarityPreservingKD,Yim2017AGF}.
% such as intermediate layers output \cite{Romero2015FitNets,Zagoruyko2017Paying,Ahn2019Variational,Tian2020ContrastiveRD} and relationships between different samples or layers \cite{Tung2019SimilarityPreservingKD,Yim2017AGF}. 
% FitNet \cite{Romero2015FitNets} utilizes feature maps in the intermediate layers as hints to further improve the final student performance, which aligns feature maps of the teacher and the student at a randomly certain layer. 
% AT \cite{Zagoruyko2017Paying} develops attention maps derived from feature maps to transfer knowledge between layers in a one-to-one manner.
% CRD \cite{Tian2020ContrastiveRD} introduces contrastive learning to capture more valuable information in the feature representation.
For example,
SP \cite{Tung2019SimilarityPreservingKD} explores the similarity between attention maps of different samples and then utilizes it to guide the student. FSP \cite{Yim2017AGF} designs a matrix to represent the relationships between feature maps of two layers, and then minimizes the difference between the student's and teacher's matrices.  
In this paper, we improve the vanilla KD technique by just employing class predictions and feature maps as the teacher knowledge to supervise shallow layers of the student model.

% Knowledge distillation is proposed to distill knowledge from a large teacher model into a small student model, serving as an model compression technique \cite{Hinton2015Distilling,chen2020online}. 
% Besides aligning final class predictions of the teacher and student models as the vanilla KD \cite{Hinton2015Distilling}, many variants utilize feature maps in the intermediate layers as the extra teacher knowledge for better knowledge transfer \cite{Romero2015FitNets,Zagoruyko2017Paying,tang2021Differentiable,Chen2021Cross,Ji2021Show,zhang2021confidence}.
% In contrast, we improve the vanilla KD technique by making shallow student layers also participate in learning the final teacher predictions. %That is to say, multiple layers of the student model are required to learn a specific layer of the teacher model.

{\bf Deep supervision.}
Deep supervision (DS) was first proposed to effectively address the convergence issue and boost the classification performance by using multiple auxiliary classifiers in shallow layers to learn ground-truth labels \cite{Lee2015Depply,Szegedy2015Going}. The significant performance improvement brought by this technique has been observed in other applications, such as objection detection \cite{Lin2017FocalLF}, semantic segmentation \cite{Zhang2018ExFuseEF} and pose estimation \cite{Newell2016StackedHN}. This technique was later applied to self-distillation \cite{Sun2019Deeply,Zhang2019BYOT}, i.e. distilling knowledge within the network itself, and online knowledge distillation \cite{Li2020OnlineKD}, i.e., training an ensemble of students collaboratively and having them teach each other.
% BYOT \cite{Zhang2019BYOT} divides the network into several blocks and then transfers knowledge from the bottom position to the shallow ones. DKS \cite{Sun2019Deeply} considers knowledge matching between different branch pairs, that is, transferring knowledge in both top-down and bottom-top directions. 
Different from these works, we use teacher knowledge rather than self-knowledge, peer knowledge or ground-truth labels as the learning target. Extensive experiments show the superiority of our strategy.

 \section{The Proposed Method}
\label{sec:method}

Given an input image $x$ with the one-hot label $y\in \mathbb{R}^{K}$, the logit output of the teacher/student model is denoted as $z_{T}/z_{S}\in \mathbb{R}^{K}$. We attach auxiliary classifiers for the $1$-th to $(L-1)$-th of the total $L$ student layers, such that these shallow student layers can learn predictions from the teacher model. The overview of our proposed method is shown in Figure~\ref{fig:overview}.

All classifiers in the student model are denoted as $C=\{c_l\}^{L}_{l=1}$, where $c_L$ is the original classifier in the last layer and the others are the newly added auxiliary classifiers in shallow layers. 
Note that these auxiliary classifiers will not increase the inference burden, since they are only utilized in the training period.
% and to enable the student model to gather different gradient information from different layers to suppress the propagation of inherent training bias, these classifiers don't share parameters.

\subsection{The Design of Auxiliary Classifiers}

\begin{table*}[t]
    \centering
     \caption{Architectural details of a WRN-16-2 as the backbone with its two auxiliary classifiers. The bracket $\left[\cdot\right]$ indicates adopting traditional convolution and the bracket  $\{\cdot\}$ indicates adopting depthwise separable convolution. }
    \begin{tabular}{c|c|c|c|c}
    \toprule
    Layer Name & Spatial Size & Backbone & $c_1$ & $c_2$ \\
    \midrule
    Stem & 32$\times$32 & 32$\times$32,16 & - & -  \\
    \midrule
    Block1 & 32$\times$32 & 
	${\begin{bmatrix}
	 3\times3,32 \\
	 3\times3,32 
	 \end{bmatrix}}\times2$ &
    - & - \\
    \midrule
    Block2 &  16$\times$16 & 
    ${\begin{bmatrix}
	 3\times3,64 \\
	 3\times3,64 
	 \end{bmatrix}}\times2$ &
	${\begin{Bmatrix}
	 3\times3,32 \\
	 1\times1,32 \\
	 3\times3,32 \\
	 1\times1,64
	 \end{Bmatrix}}\times1$ & - \\
	 \midrule
	  Block3 &  8$\times$8 & 
    ${\begin{bmatrix}
	 3\times3,128 \\
	 3\times3,128 
	 \end{bmatrix}}\times2$ &
	${\begin{Bmatrix}
	 3\times3,64 \\
	 1\times1,64 \\
	 3\times3,64 \\
	 1\times1,128
	 \end{Bmatrix}}\times1$ & 
	 ${\begin{Bmatrix}
	 3\times3,64 \\
	 1\times1,64 \\
	 3\times3,64 \\
	 1\times1,128
	 \end{Bmatrix}}\times1$ \\
	 \midrule
	 Pooling & 1$\times$1 & \multicolumn{3}{c}{Global Average Pooling} \\
	 \midrule
	 Linear Classifier & 1$\times$1 & \multicolumn{3}{c}{Fully-Connected layer} \\
	 \bottomrule

    \end{tabular}
   
    \label{table:aux}
\end{table*}

Generally, convolutional neural networks will downsample feature maps along the spatial dimension to capture semantic information of different granularities \cite{Chen2019DropAO}. High-resolution feature maps from shallow layers usually extract fine-grained details and high frequency patterns, such as corners, edges and textures, while low-resolution feature maps from deep layers contain coarse-grained information for better classification performance, such as faces, legs and global structures \cite{Huang2018Multi,Zeiler2014VisualizingAU}.

Coarse-grained features are important for classifying the content of the whole image.
Therefore, adding a simple auxiliary classifier (e.g., a fully-connected layer) to shallow layers is not suitable due to the lack of coarse-grained feature, which may even hurt final prediction performance \cite{Huang2018Multi}.
To address this issue, we design a complex auxiliary classifier, which consists of multiple convolution blocks, a global average pooling layer and a fully-connected layer, after feature maps of shallow layers to obtain such coarse-grained features. 
For reducing computational burden, we adopt depthwise separable convolution \cite{Chollet2017XceptionDL} to build lightweight classifiers. 

% We make each block of the auxiliary classifier have
% the same downsampling  number as the corresponding block of the backbone for appropriate fine-to-coarse feature transformation.

As shown in Table~\ref{table:aux}, we take WRN-16-2 as an instance \cite{Zagoruyko2016Wide}, which contains three block layers, to systematically display architectural details of the backbone and its two auxiliary classifiers. Considering that the overly complex classifier may cause the same training bias accumulation problem happen again, we try to design the auxiliary classifier as simple as possible under the premise of guaranteeing coarse-grained features. The design principle of the auxiliary classifier is to ensure that it has the same downsampling path as the main branch to imitate a similar fine-to-coarse feature transformation procedure. So for lightweight and broad applicability, the number of auxiliary classifier blocks is set to 1 in any network, which is sufficient to satisfy the design principle.

% Since shallow layers only contain fine-level features but lack of coarse-level features, they are not suitable to be directly used with a fully-connected layer for the final prediction \cite{Huang2018Multi}. Additional convolutional layers are generally needed to obtain such coarse-level features.For simplicity and broad applicability, we design the architecture of each auxiliary classifier to be the one that helps the corresponding shallow layer same as the main branch.

\begin{table*}[t]
  \centering
  \caption{Top-1 test accuracy of
knowledge distillation methods on CIFAR-100.}
  \label{table:compare_kd_cifar100}
  \resizebox{0.98\textwidth}{!}{
  \begin{tabular}{c|ccccccc}
  \toprule
  \multirow{2}*{Teacher}&WRN-40-2 &VGG13 &ResNet32x4 &WRN-40-2 &ResNet32x4 &ResNet32x4 &ResNet32x4 \\
  &75.61 &74.64 &79.42 &75.61 &79.42 &79.42 &79.42 \\
  \midrule
  \multirow{2}*{Student}&WRN-40-1 &VGG8 &ResNet8x4 &WRN-16-2 &VGG8 &ShuffleNetV2 &MobileNetV2 \\
  &72.06 &70.51 &73.05 &73.13 &70.51 &72.99 &65.41  \\
  \midrule
  KD \cite{Hinton2015Distilling}   &74.05$\pm$0.11 &73.34$\pm$0.04  &74.39$\pm$0.21 &75.37$\pm$0.17 &72.58$\pm$0.26 &75.55$\pm$0.05 &67.31$\pm$0.18 \\
  FitNet \cite{Romero2015FitNets} &74.20$\pm$0.04 &73.35$\pm$0.30 &74.26$\pm$0.15 &75.17$\pm$0.21 &72.98$\pm$0.10 &75.62$\pm$0.20 &64.46$\pm$0.34 \\
  AT \cite{Zagoruyko2017Paying}   &74.06$\pm$0.38 &73.65$\pm$0.16 &75.11$\pm$0.04 &75.66$\pm$0.06 &71.96$\pm$0.06 &76.03$\pm$0.12 &67.17$\pm$0.10\\
  VID \cite{Ahn2019Variational} & 73.80$\pm$0.14 & 73.67$\pm$0.04 & 74.57$\pm$0.03 & 75.25$\pm$0.03 & 73.33$\pm$0.14 & 75.85$\pm$0.30 & 67.98$\pm$0.22 \\
  CRD \cite{Ahn2019Variational} &74.23$\pm$0.13 &74.28$\pm$0.03  &75.70$\pm$0.12 &75.82$\pm$0.13 &73.72$\pm$0.28 &76.48$\pm$0.04 &69.01$\pm$0.11 \\
 ICKD \cite{Liu2021ExploringIC}  &74.36$\pm$0.18 &73.65$\pm$0.23  &74.91$\pm$0.56 &75.44$\pm$0.12 &73.52$\pm$0.14 &75.76$\pm$0.33 &67.82$\pm$0.25 \\
 DIST \cite{Huang2022KnowledgeDF} & 74.73$\pm$0.24 & 73.90$\pm$0.35 & 76.31$\pm$0.19 &  75.66$\pm$0.11 & 73.70$\pm$0.05 & 77.35$\pm$0.25 & 68.01$\pm$0.01 \\
  \midrule
    DSKD &\bf75.29$\pm$0.18 &\bf74.40$\pm$0.05 &\bf76.43$\pm$0.04 &\bf76.50$\pm$0.15 &\bf75.01$\pm$0.02 &\bf78.05$\pm$0.03 &\bf69.42$\pm$0.13 \\
  \bottomrule
  \end{tabular}}
%  \vspace{-0.3cm}
  \end{table*}

\subsection{The Loss of Class Predictions}
\label{subsec:loss_pred}
As for the vanilla knowledge distillation  \cite{Hinton2015Distilling}, class predictions of the teacher and student models are required to be aligned in the last layer. 
The associated loss is defined as the Kullback-Leibler (KL) divergence between the teacher output $z_T$ and the output of the last student classifier $c_L$, i.e., $z_{c_L}$
\begin{equation}
  \mathcal L_{{KD}_{last}}=KL\left(\sigma\left(z_T/\tau\right)||\sigma\left(z_{c_L}/\tau\right)\right),
\end{equation}
where $\sigma(\cdot)$ is a softmax function and temperature $\tau$ is a hyper-parameter. A higher $\tau$ makes the distribution softer.

We generalize this technique by involving shallow layers in the learning of teacher class predictions. In this way, the student model gathers gradient information not only from the last layer but also from those shallow layers to suppress the propagation of training bias. 
The shallow layer loss is defined as
\begin{equation}
  \mathcal L_{{KD}_{shallow}}=\sum_{l=1}^{L-1}{W_{KD}^{c_l}KL\left(\sigma\left(z_T/\tau\right)||\sigma\left(z_{c_l}/\tau\right)\right)},
  \label{equation 2}
\end{equation}
where $W_{KD}^{c_l}$ is an adaptive weight for the training of the auxiliary classifier $c_l$. We will elaborate this in Section~\ref{subsec:weight}.

The total loss of class predictions is summarized as
\begin{equation}
  \mathcal L_{KD}=\mathcal L_{{KD}_{shallow}}+\mathcal L_{{KD}_{last}}.
\end{equation}

\subsection{The Loss of Feature Maps}
Besides class predictions, feature maps can also help improve the student model performance \cite{Romero2015FitNets,Zagoruyko2017Paying,Ahn2019Variational}. Thus, we take feature maps of the last teacher layer $F_T$ as another learning target. 
The feature maps generated by all student classifiers (before the global average pooling layer) are denoted as
$F_{c_1}, F_{c_2},…, F_{c_L}$, respectively.

Similar to the discussion in Section \ref{subsec:loss_pred}, we calculate the Mean-Square-error (MSE) loss between the teacher feature maps $F_T$ and feature maps in the 
last student layer $F_{c_L}$ as 
\begin{equation}
  \mathcal L_{{Fea}_{last}}=MSE\left(F_T,r\left(F_{c_L}\right)\right),
  \label{eq:fea_last}
\end{equation}
where $r(\cdot)$ is a projection function to make dimensions of feature maps to be aligned. 

We then generalize the above loss function by involving shallow layers in the learning of teacher feature maps and define the shallow layer loss as follows
\begin{equation}
  \mathcal L_{{Fea}_{shallow}}=\sum_{l=1}^{L-1}{W_{Fea}^{c_l}MSE\left(F_T,r\left(F_{c_l}\right)\right)},
  \label{equation 5}
\end{equation}
where $W_{Fea}^{c_l}$ is an adaptive weight for the training of the auxiliary classifier $c_l$.  We will elaborate this  in Section~\ref{subsec:weight}.

Note that the last teacher layer feature is used to supervise the last layer feature of each shallow auxiliary classifier, rather than intermediate features of the student directly. In this way, the feature of each shallow classifier and that of the teacher is corresponding. Furthermore, the last layer feature contains coarse-grained features, which are important for classifying the content of the whole image into a single class \cite{Huang2018Multi}. So it is more reliable to utilize the last feature layer as supervision.

The total loss of feature maps is summarized as
\begin{equation}
  \mathcal L_{Fea}=\mathcal L_{{Fea}_{shallow}}+\mathcal L_{{Fea}_{last}}.
  \label{equation 6}
\end{equation}

\subsection{Loss-based Weights}
\label{subsec:weight}

% different initialization of shallow layers will cause differences in the final prediction results. 
Each shallow layer classifier would show a different behaviour in the training process due to the different initialization \cite{Skorski2021Revisiting}. 
If each classifier is simply assigned with an average weight, the final model performance would be negatively affected from those classifiers falling behind in a certain iteration. We thus develop a loss-based weighting strategy, which measures the confidence of each auxiliary classifier on each sample, to alleviate this effect. The formulation is defined as 
\begin{equation}
  W_{KD}^{c_l}=\frac{KL\left(\sigma\left(z_T/\tau\right)||\sigma\left(z_{c_l}/\tau\right)\right)}{\sum_{j=1}^{L-1}{KL\left(\sigma\left(z_T/\tau\right)||\sigma\left(z_{c_j}/\tau\right)\right)}},
  \label{equation 9}
\end{equation}
\begin{equation}
  W_{Fea}^{c_l}=\frac{MSE(F_T,r\left(F_{c_l}\right))}{\sum_{j=1}^{L-1}{MSE(F_T,r(F_{c_j}))}}.
  \label{equation 10}
\end{equation}
A larger weight is allocated to the auxiliary classifier with a larger loss value to make it catch up with the training process.

% Only when the process of each auxiliary classifier reach a balance can the overall early learning effectiveness be maximized.

\subsection{The Overall Loss Function}
%\begin{equation}
%  \mathcal L_{CE}=CrossEntropy\left(y,\sigma\left(z_{c_L}\right)\right),
%\end{equation}
The final total loss is summarized as
\begin{equation}
  \mathcal L_{Total}=\mathcal L_{CE}+\alpha \mathcal L_{KD}+\beta \mathcal L_{Fea},
\end{equation}
where $\alpha$ and $\beta$ are hyper-parameters utilized to balance three loss items. $\mathcal{L}_{CE}=CrossEntropy\left(y,\sigma\left(z_{c_L}\right)\right)$ is a standard cross-entropy loss calculated between class predictions and labels in the classification task.

\section{Experiment}
\label{sec:experiments}

We conduct all experiments on two public image classification datasets: CIFAR-100 \cite{krizhevsky2009learning} and TinyImageNet \cite{Le2015TinyIV}. To demonstrate the effectiveness of our proposed DSKD, we use seven groups of teacher-student models with different networks covering VGG \cite{simonyan2015Very}, ResNet \cite{He2016Deep}, WRN \cite{Zagoruyko2016Wide}, MobileNet \cite{Sandler2018MobileNetV2} and ShuffleNet \cite{zhang2018shufflenet,Ma2018shufflenv2}. 
We first compare with several representative knowledge distillation methods \cite{Hinton2015Distilling,Romero2015FitNets,Zagoruyko2017Paying,Ahn2019Variational,Tian2020ContrastiveRD,Liu2021ExploringIC} and deep supervision methods \cite{Lee2015Depply,Zhang2019BYOT,Sun2019Deeply,yao2020knowledge} and then analyze the impact of shallow layer loss. we also perform ablation experiments to verify the effectiveness of each module in our DSKD and visualize the distribution of learning weights. Finally, we analyze the sensitivity of the hyper-parameter $\beta$ and the computational burden of our proposed method.

\subsection{ Datasets and Settings}

{\bf Datasets.} There are two datasets in our experiments. CIFAR-100 consists of 60000 colored images from 100 categories. Each category has 600 images, of which 500 are used as training set and 100 are used as test set. TinyImageNet is a more challenging dataset, which is a subset of the ILSVRC-2012 dataset and consists of 200 categories. In each category, there are 500 images for training, 50 images for validation and 50 images for test.

{\bf Settings.} We implement all methods in Pytorch \cite{Paszke2019PyTorchAI} and conduct all experiments on an NVIDIA GeFore RTX 2080Ti GPU. For all datasets, we train all models for 240 epochs with a batch size of 64 and the learning rate is divided by 10 at 150th, 180th and 210th epochs. The initial learning rate is 0.01 for MobileNet and ShuffleNet, and 0.05 for other models. The weight decay is set to $5\times{10}^{-4}$. For fairness of comparison, we set $\alpha$ to 1 and temperature $\tau$ to 4 for all methods. The hyper-parameter $\beta$ in our proposed DSKD is set to 30. To ensure the reliability of the results, we train each method for three times and report the means and standard deviations.

\begin{figure*}[t]
  \centering
  \includegraphics[width=0.95\textwidth]{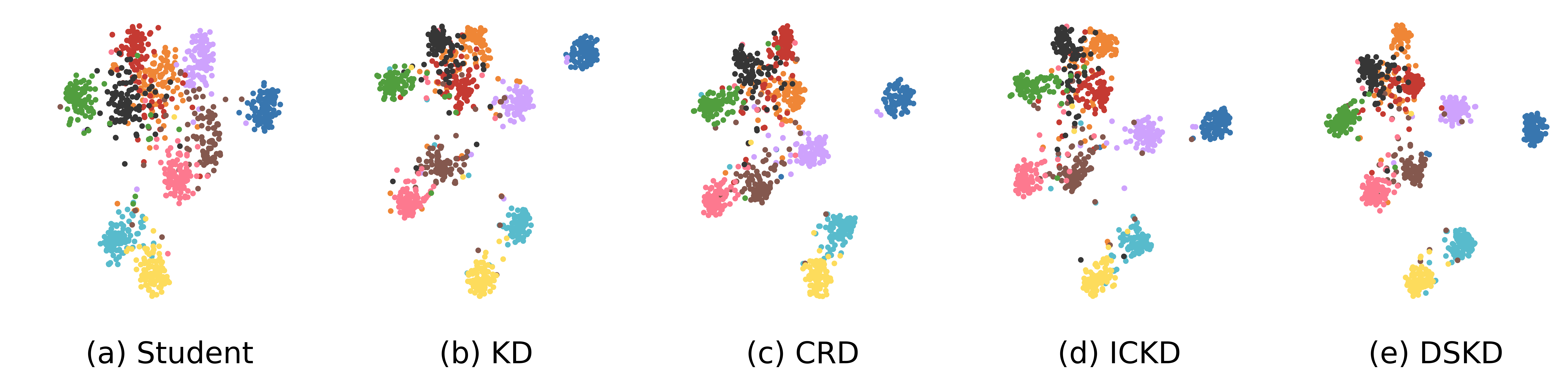}
  \caption{t-SNE visualization of feature distribution on CIFAR-100.}
  \label{fig:tsne}
% \vspace{-0.3cm}
\end{figure*}

\subsection{Comparison with Knowledge Distillation Methods}

We compare with seven popular knowledge distillation methods on seven groups of teacher-student models, including four homogeneous and three heterogeneous architecture combinations. The compared methods are as follows:
\begin{itemize}
\item Vanilla KD \cite{Hinton2015Distilling}: the original KD method that matches class predictions between the last layer of the teacher and the student.
\item FitNet \cite{Romero2015FitNets}: a feature-based KD method that minimizes the distance between the feature maps of a selected teacher-student layer pair.
\item AT \cite{Zagoruyko2017Paying}: a feature-based KD method that minimizes the distance between the attention maps of all teacher-student layer pairs.
\item VID \cite{Ahn2019Variational}: a feature-based KD method that maximizes mutual information between the teacher and the student networks.
\item CRD \cite{Tian2020ContrastiveRD}: a feature-based KD method that introduces contrastive learning objective to maximize the mutual information between feature representations of the teacher and the student.
\item ICKD \cite{Liu2021ExploringIC}: a feature-based KD method that minimizes the distance between the teacher and the student Inter-Channel Correlation matrix calculated from feature maps.
\item DIST \cite{Huang2022KnowledgeDF}: a correlation-based KD method that minimizes Pearson’s distance between class predictions of the teacher and the student.
\end{itemize}
% KD \cite{Hinton2015Distilling} uses class predictions as the teacher knowledge, while AT \cite{Zagoruyko2017Paying}, FitNet \cite{Romero2015FitNets} and VID \cite{Ahn2019Variational} use intermediate feature maps to supervise the student model. 

% \begin{table}
%     \centering
%     \begin{tabular}{c|c|c|c|c|c|c|c|c}
%     \hline
%     \multirow{2}{*}{Teacher} & \multirow{2}{*}{Student} & \multicolumn{7}{c}{Test accuracy} \\
%     \cline{3-9}
%     & & Teacher & Student & KD \cite{Hinton2015Distilling} & FitNet \cite{Romero2015FitNets} & AT \cite{Zagoruyko2017Paying} & CRD \cite{Tian2020ContrastiveRD} & DSKD \\
%     \hline
%     WRN-40-2 & WRN-40-1 & 75.61\% & 72.06\% & 74.05\% & 74.20\% & 74.06\% & 74.23\% & \bf75.07\% \\
%     \hline
%     VGG13 & VGG8 & 74.64\% & 70.51\%  &73.34\% & 73.35\% & 73.65\% & 74.28\% & \bf74.55\% \\
%     \hline
%     Resnet32x4 & Resnet8x4 & 79.42\% & 73.05\% &74.39\% & 74.26\% & 75.11\% & 75.70\% & \bf76.31\%  \\
%     \hline
%     WRN-40-2 & WRN-16-2 & 75.61\% & 73.13\% & 75.37\% & 75.17\% & 75.66\% & 75.82\% & \bf76.50\% \\
%     \hline
%     ResNet32x4 & VGG8 & 79.42\% & 70.51\% & 72.58\% & 72.98\% & 71.96\% & 73.72\% & \bf75.04\% \\
%     \hline
%     ResNet32x4 & ShuffleNetV2 & 79.42\% & 72.99\% & 75.55\% & 75.62\% & 76.03\% & 76.48\% & \bf77.51\% \\
%     \hline
%     WRN-40-2 & MobileNetV2 & 75.61\% & 65.41\% & 68.62\% & 67.95\% & 68.88\% & 69.01\% & \bf69.42\% \\
%     \hline
%     \end{tabular}
%     \caption{Top-1 test accuracy of knowledge distillation methods comparison on CIFAR-100 dataset.}
%     \label{table: compare_kd_cifar100}
% \end{table}

{\bf Results on CIFAR-100.} 
% Table~\ref{table:compare_kd_cifar100} shows top-1 test accuracy of knowledge distillation methods on CIFAR-100. From the results, 
As shown in Table~\ref{table:compare_kd_cifar100}, our proposed method consistently outperforms all other methods.  Our method exceeds KD, FitNet, AT, VID, CRD, ICKD and DIST with 1.79\%, 2.11\%, 1.62\%, 1.52\%, 0.84\%, 1.24\% and 0.78\% average improvement \footnote{average  improvement=$\frac{1}{N}\sum_{i=1}^{N}(Acc_{DSKD}^{i}-ACC_{COMP}^{i})$, where $Acc_{DSKD}^{i}$ and $ACC_{COMP}^{i}$ refer to the accuracy of our DSKD and a compared method in the $i^{th}$ teacher-student combination.} respectively. We also observe that the performance of FitNet, AT and VID are similar to KD.

As shown in Fig.~\ref{fig:tsne}, experiments on CIFAR-100 for Resnet32x4 \& ShuffleNetV2 are conducted to compare feature distribution of student networks under different training methods. We randomly select 10 out of 100 classes and visualize feature distribution with t-SNE \cite{van2008visualizing}. We can find that the features from the student trained with KD are more distinguishable between different classes than the student trained from scratch, which means teacher knowledge can effectively improve discrimination ability of the student network. Compared to other KD methods, our DSKD makes the features more compact among the same classes, such as the brown, orange and 
green clusters in the figure, which demonstrates the superiority of our method. 

% In order to intuitively quantify the improvement of our method, following the previous work \cite{Tian2020ContrastiveRD}, we employ an evaluation indicator called Average Relative Improvement (ARI): 
% \begin{equation}
%     ARI=\frac{1}{N}\sum
% \end{equation}
% The comparison results in Table~\ref{table:compare} show that our proposed method consistently outperforms all other methods. For example, our DSKD achieves about 1.46\% absolute accuracy improvement over VID \cite{Ahn2019Variational}, averaged across these seven teacher-student combinations.

% \begin{table}
%     \centering
%     \begin{tabular}{c|c|c|c|c|c|c|c|c}
%     \hline
%     \multirow{2}{*}{Teacher} & \multirow{2}{*}{Student} & \multicolumn{7}{c}{Test accuracy} \\
%     \cline{3-9}
%     & & Teacher & Student & KD & FitNet & AT & CRD & DSKD \\
%     \hline
%     WRN-40-2 & WRN-40-1 & 56.35\% & 48.14\% & 51.12\% & 50.83\% & 51.44\% & 45.05\% & \bf52.12\% \\
%     \hline
%     VGG13 & VGG8 & 54.39\% & 48.84\% & 52.82\% & 52.61\% & 51.05\% & 47.77\% & \bf53.15\% \\
%     \hline
%     ResNet32x4 & VGG8 & 60.56\% & 48.84\% & 53.18\% & 52.69\% & 51.64\% & 49.07\% & \bf54.47\% \\
%     \hline
%     ResNet32x4 & ShuffleNetV2 & 60.56\% & 55.47\% & 58.37\% & 58.27\% & 57.88\% & 53.37\% & \bf60.25\% \\
%     \hline
%     \end{tabular}
%     \caption{Top-1 test accuracy of knowledge distillation methods comparison on TinyImageNet.}
%     \label{table: compare_kd_tiny}
% \end{table}

\begin{table}[t]
  \centering
  \caption{Top-1 test accuracy of
knowledge distillation methods on TinyImageNet.}
  \label{table: compare_kd_tiny}
  %\resizebox{0.98\textwidth}{!}{
  \begin{tabular}{c|cccc}
  \toprule
  \multirow{2}*{Teacher}&WRN-40-2 &VGG13 &ResNet32x4  &ResNet32x4  \\
  & 56.35 & 54.39 & 60.56 & 60.56 \\
  \midrule
  \multirow{2}*{Student}&WRN-40-1 &VGG8 &VGG8 &ShuffleNetV2 \\
  & 48.14 & 48.84 & 48.84 & 55.47  \\
  \midrule
  KD \cite{Hinton2015Distilling} & 51.12$\pm$0.02 & 52.82$\pm$0.06 & 53.18$\pm$0.02 & 58.37$\pm$0.15    \\
  FitNet \cite{Romero2015FitNets} & 50.83$\pm$0.10 & 52.61$\pm$0.10 & 52.69$\pm$0.17 & 58.27$\pm$0.23 \\
  AT \cite{Zagoruyko2017Paying} & 51.44$\pm$0.20 & 51.05$\pm$0.42 & 51.64$\pm$0.27 & 57.88$\pm$0.21  \\
  VID \cite{Ahn2019Variational} & 47.71$\pm$0.22 & 47.19$\pm$0.18 & 48.73$\pm$0.45 & 53.94$\pm$0.29 \\
  CRD \cite{Tian2020ContrastiveRD} & 45.05$\pm$0.15 & 47.77$\pm$0.44 & 49.07$\pm$0.08 & 53.37$\pm$0.22  \\
  ICKD \cite{Liu2021ExploringIC} & 51.26$\pm$0.16 & 52.08$\pm$0.15 & 53.30$\pm$0.24 & 59.07$\pm$0.13 \\ 
  DIST \cite{Huang2022KnowledgeDF} &51.63$\pm$0.21 &  \bf53.80$\pm$0.10 & 54.01$\pm$0.15 & 59.80$\pm$0.20 \\
  \midrule
  DSKD & \bf52.12$\pm$0.30 & 53.15$\pm$0.30 & \bf54.47$\pm$0.23 & \bf60.25$\pm$0.12  \\
  \bottomrule
  \end{tabular}%}
%  \vspace{-0.3cm}
  \end{table}

{\bf Results on TinyImageNet.}  As shown in Table~\ref{table: compare_kd_tiny},  our proposed method also outperforms most other methods on TinyImageNet which is a more challenging dataset. Our method exceeds KD, FitNet, AT, VID, CRD, ICKD and DIST with 1.13\%, 1.39\%, 1.20\%, 5.6\%, 6.18\%, 1.07\% and 0.19\% average improvement respectively. We even observe an incredible result that the third-best method on CIFAR-100 (CRD) turns out to be the worst on TinyImageNet, while our method consistently maintains good performance. These results show that our method is more effective and stable than other methods.
\begin{table*}[t]
  \centering
  \caption{Top-1 test accuracy of deep supervision methods on CIFAR-100.}
  \label{table: compare_deeply_cifar100}
  \resizebox{0.98\textwidth}{!}{
  \begin{tabular}{c|ccccccc}
  \toprule
  \multirow{2}*{Teacher}&WRN-40-2  &ResNet32x4 &WRN-40-2 &ResNet32x4 &ResNet32x4 &ResNet32x4 &VGG13 \\
  &75.61  &79.42 &75.61 &79.42 &79.42 &79.42 & 74.64 \\
  \midrule
  \multirow{2}*{Student}&WRN-40-1  &ResNet8x4 &WRN-16-2 &VGG8 &ShuffleNetV2 &MobileNetV2 &ShuffleNetV1 \\
  &72.06  &73.05 &73.13 &70.51 &72.99 &65.41 &71.36   \\
  \midrule
  DSN \cite{Lee2015Depply}     &72.63$\pm$0.02 &73.10$\pm$0.09 &74.25$\pm$0.10 &71.26$\pm$0.44 &73.98$\pm$0.12 &65.76$\pm$0.03 & 72.14$\pm$0.12 \\
  BYOT \cite{Zhang2019BYOT}    &72.37$\pm$0.35 &72.98$\pm$0.04 &73.70$\pm$0.13 &70.88$\pm$0.17 &74.32$\pm$0.05 & 64.93$\pm$0.65 & 72.77$\pm$0.08  \\
  DKS \cite{Sun2019Deeply} & 73.43$\pm$0.19 &73.51$\pm$0.07 & 74.68$\pm$0.06 &72.01$\pm$0.06 & 75.12$\pm$0.13 & 66.26$\pm$0.15 & 73.33$\pm$0.23 \\
  DCM \cite{yao2020knowledge} & 72.91\%$\pm$0.13 &  74.75$\pm$0.28 & 75.31$\pm$0.18 & 73.00$\pm$0.24 & 76.22$\pm$0.39 & 67.07$\pm$0.25 & 74.71$\pm$0.05 \\
  \midrule
   DSKD &\bf75.29$\pm$0.18 &\bf76.43$\pm$0.04 &\bf76.50$\pm$0.15 &\bf75.01$\pm$0.02 &\bf78.05$\pm$0.03 &\bf69.42$\pm$0.13 &\bf75.70$\pm$0.13 \\
  \bottomrule
  \end{tabular}}
%  \vspace{-0.3cm}
  \end{table*}

  \begin{table}[t]
  \centering
  \caption{Top-1 test accuracy of
deep supervision methods on TinyImageNet.}
  \label{table: compare_deeply_tiny}
  \resizebox{1\linewidth}{!}{
  \begin{tabular}{c|cccc}
  \toprule
  \multirow{2}*{Teacher}&WRN-40-2 &ResNet32x4  &ResNet32x4 &VGG13  \\
  & 56.35 & 60.56 & 60.56 &54.39 \\
  \midrule
  \multirow{2}*{Student}&WRN-40-1&VGG8 &ShuffleNetV2 &ShuffleNetV1 \\
  & 48.14 & 48.84 & 55.47 & 53.85 \\
  \midrule
  DSN \cite{Lee2015Depply} & 49.86$\pm$0.23 & 47.73$\pm$0.25 & 52.7$\pm$0.11 &54.51$\pm$0.32   \\
  BYOT \cite{Zhang2019BYOT} & 48.93$\pm$0.11 & 46.63$\pm$0.08 & 53.27$\pm$0.23 &54.19$\pm$0.12 \\
  DKS \cite{Sun2019Deeply} & 50.59$\pm$0.22 & 48.46$\pm$0.11 & 54.81$\pm$0.23 &55.73$\pm$0.02 \\
  DCM \cite{yao2020knowledge} & 47.41$\pm$0.11 & 51.00$\pm$0.14 & 57.39$\pm$0.22 & 55.85$\pm$0.14\\
  \midrule
  DSKD & \bf52.12$\pm$0.30 & \bf54.47$\pm$0.23 & \bf60.25$\pm$0.12 & \bf56.17$\pm$0.05 \\
  \bottomrule
  \end{tabular}}
%  \vspace{-0.3cm}
  \end{table}

\subsection{Comparison with Deep Supervision Methods}

We also compare with four popular deep supervision methods that give supervisory signals for shallow layers similarly.The compared methods are as follows:
\begin{itemize}
    \item DSN \cite{Lee2015Depply}: the original DS method that takes ground-truth labels as supervisory signals for shallow layers.
    \item BYOT \cite{Zhang2019BYOT}: a self-distillation DS method that mainly makes shallow layers learn class predictions and feature maps of the student's own last layer. 
    \item DKS \cite{Sun2019Deeply}: a self-distillation DS method that enable each layer pair to learn class predictions from each other.
    \item DCM \cite{yao2020knowledge}: an online distillation DS method that utilizes class predictions of a peer network to supervise shallow layers by dense cross-layer transfer.
\end{itemize}

% DSN \cite{Lee2015Depply} uses ground-truth labels while BYOT \cite{Zhang2019BYOT} and DKS \cite{Sun2019Deeply} use outputs of the student own layer as supervisory signals. Different from them, we use predictions from a pre-trained teacher model as supervisory signals.

{\bf Results on CIFAR-100.} As shown in Table~\ref{table: compare_deeply_cifar100}, our method outperforms all other methods by a large margin. Our method exceeds DSN, BYOT, DKS and DCM with 3.33\%, 3.49\%, 2.58\% and 1.77\% average improvement respectively. We believe that teacher knowledge provides better guidance to improve the student generalization ability than ground-truth labels (compared to DSN and verified in section~\ref{sec:ablation}), student's own knowledge (compared to BYOT and DKS) and peer knowledge (compared to DCM). 

% \begin{table}
%     \centering
%     \begin{tabular}{c|c|c|c|c|c|c|c}
%     \hline
%     \multirow{2}{*}{Teacher} & \multirow{2}{*}{Student} & \multicolumn{6}{c}{Test accuracy} \\
%     \cline{3-8}
%     & & Teacher & Student & DSN & BYOT & DKS & DSKD \\
%     \hline
%     WRN-40-2 & WRN-40-1 & 56.35\% & 48.14\% & 49.86\% & 48.93\% & 50.59\%  & \bf52.12\% \\
%     \hline
%     ResNet32x4 & VGG8 & 60.56\% & 48.84\% & 47.73\% & 46.63\% & 48.46\% &  \bf54.47\% \\
%     \hline
%     ResNet32x4 & ShuffleNetV2 & 60.56\% & 55.47\% & 52.7\% & 53.27\% & 54.81\% & \bf60.25\% \\
%     \hline
%     \end{tabular}
%     \caption{Top-1 test accuracy of knowledge distillation methods comparison on TinyImageNet.}
%     \label{table: compare_deeply_tiny}
% \end{table}

\begin{table}[t]
    \centering
    \caption{Impact of the shallow layer loss $\mathcal{L}_{{KD}_{shallow}}$ and $\mathcal{L}_{{Fea}_{shallow}}$ on CIFAR-100 for ResNet32x4 \& ShuffleNetV2.}
    \begin{tabular}{cc|cc|c}
     \toprule
     Layer-1 & Layer-2 & $\mathcal{L}_{{KD}_{shallow}}$ & $\mathcal{L}_{{Fea}_{shallow}}$ & Test accuracy \\
     \midrule
      
         &  & & & 76.89$\pm$0.12 \\
    \checkmark& & \checkmark &  & 77.22$\pm$0.11 \\
     \checkmark& &  & \checkmark & 77.20$\pm$0.04 \\
    \checkmark& & \checkmark & \checkmark & 77.40$\pm$0.12 \\
     & \checkmark & \checkmark &  & 77.57$\pm$0.13 \\
      & \checkmark &  & \checkmark & 77.07$\pm$0.21 \\
         & \checkmark & \checkmark & \checkmark & 77.78$\pm$0.04 \\
    
      \checkmark & \checkmark & \checkmark & \checkmark &  \bf78.05$\pm$0.03 \\
     \bottomrule
    \end{tabular}
    
    \label{table:impact}
\end{table}

{\bf Results on TinyImageNet.} As shown in Table~\ref{table: compare_deeply_tiny}, our method also outperforms all other methods by a larger margin on such a challenging dataset.  Our method exceeds DSN, BYOT, DKS and DCM with 4.55\%, 5.00\%, 3.36\% and 2.84\% average improvement respectively. 
We even observe that all competitors sometimes performs worse than the student model itself. This indicates that the knowledge learned in the early student training period may be very noisy, and supplying it for shallow layers would result in a negative impact on the final performance, and hard labels, i.e., ground-truth labels, may inhibit generalization  ability of the student.

% Our DSKD achieves considerable performance improvement over these previous works. Experimental results in Table~\ref{table:compare} show that teacher knowledge provides better guidance to improve the student generalization ability than ground-truth labels (compared to DSN \cite{Lee2015Depply} and verified in section~\ref{sec:ablation}) and student's own knowledge (compared to BYOT \cite{Zhang2019BYOT} and DKS \cite{Sun2019Deeply}). 
% Additionally, we also find that BYOT sometimes performs worse than the student model itself. This indicates that the knowledge learned in the early student training period may be very noisy, and supplying it for shallow layers would result in a negative impact on the final performance.

\subsection{Impact of Shallow Layer Loss}

% \begin{table}[t]
%     \centering
%     \caption{Impact of the shallow layer loss $\mathcal{L}_{{KD}_{shallow}}$ and $\mathcal{L}_{{Fea}_{shallow}}$ on CIFAR-100 for ResNet32x4 \& ShuffleNetV2.}
%     \begin{tabular}{c|c|c}
%      \toprule
%      Layer block-1 & Layer block-2 & Test accuracy \\
%      \midrule
%          &  & 76.89\% \\
%     \midrule
%     \checkmark & & 77.4\% \\
%      \midrule
%          & \checkmark & 77.78\% \\
%     \midrule
%       \checkmark & \checkmark &  \bf78.05\% \\
%      \midrule
%     \end{tabular}
    
%     \label{table:impact}
% \end{table}

We further take ``ResNet32x4 \& ShuffleNetV2" as an example to verify the impact of our proposed shallow layer loss.
Since the ShuffleNetV2 model for CIFAR-100 contains three building blocks \cite{Ma2018shufflenv2}, we treat 
outputs of the first two blocks as the possible position for adding our shallow layer loss $\mathcal{L}_{{KD}_{shallow}}$ and $\mathcal{L}_{{Fea}_{shallow}}$. 

As shown in Table~\ref{table:impact}, adding our proposed shallow layer loss effectively improves model performance. 
%Multiple layers are better than a single layer, and the earlier position is better.
In the case of no shallow layer loss is employed (the second column), the student model accuracy (76.89\%) is still better than the KD counterpart (75.55\%), which is credited to the extra feature maps loss in Equation~(\ref{eq:fea_last}).

\begin{table}[t]
    \centering
      \caption{Ablation study on CIFAR-100 for ResNet32x4 \& ShuffleNetV2.}
    \begin{tabular}{cc|c}
     \toprule
     $\mathcal{L}_{Fea}$ & Adaptive weights & Test accuracy \\
     \midrule
         &  & 76.52$\pm$0.05 \\
     & \checkmark & 76.70$\pm$0.02 \\
       \checkmark  &  &  77.39$\pm$0.02 \\
      \checkmark & \checkmark &  \bf78.05$\pm$0.03 \\
     \bottomrule
    \end{tabular}
  
    \label{table:ablation}
\end{table}

\begin{table}[t]
    \centering
     \caption{Impact of the structure of the auxiliary classifier on CIFAR-100 for ResNet32x4 \& ShuffleNetV2}
    \begin{tabular}{c|c}
    \toprule
        Method & Test Accuracy \\
    \midrule
         baseline& 76.89$\pm$0.12 \\
         with simple auxiliary classifiers & 76.93$\pm$0.19 \\
         with complex auxiliary classifiers & \bf78.05$\pm$0.03 \\
    \bottomrule
    \end{tabular}
   
    \label{table:aux_impact}
\end{table}

% \begin{table}
% \centering
% \caption{Impact of the shallow layer loss $\mathcal{L}_{{KD}_{shallow}}$ and $\mathcal{L}_{{Fea}_{shallow}}$ with ResNet32x4 \& ResNet8x4.}
% \resizebox{0.98\columnwidth}{!}{
% \begin{tabular}{c|cccc}
% \hline\noalign{\smallskip}
% Layer block-1 & &\checkmark& & \checkmark\\
% Layer block-2 & & &\checkmark &\checkmark\\
% \hline\noalign{\smallskip}
% Test accuracy & 74.98 $\pm$0.04 & 76.23$\pm$0.03 &75.97$\pm$0.03 &\bf76.31$\pm$0.04\\
% \hline\noalign{\smallskip}
% \end{tabular}}
% \label{table:impact}
% \end{table}

\subsection{Ablation Study}
\label{sec:ablation}

% \begin{table}[t]
% \centering
% \caption{Ablation study on CIFAR-100 for ResNet32x4 \& ShuffleNetV2.} 
% \resizebox{0.98\columnwidth}{!}{
% \begin{tabular}{c|cccc}
% \toprule
% $\mathcal{L}_{Fea}$
% & & & \checkmark & \checkmark\\
% Adaptive weights & & \checkmark &  & \checkmark \\
% \midrule
% Test accuracy & 76.52\% & 76.70\% & 77.39\% & \bf 78.05\% \\
% \bottomrule
% \end{tabular}}
% \label{table:ablation}
% %\vspace{-0.3cm}
% \end{table}

As shown in Table~\ref{table:ablation}, removing $\mathcal{L}_{Fea}$ or simply assigning equal weights to each shallow layer for each sample causes a considerable drop in accuracy, which demonstrates the importance of our used feature-based knowledge and loss-based weight allocation strategy.
% each module in DSKD. 
Note that even if these two modules are both removed, i.e., we only use class predictions of the teacher to supervise shallow layers of the student model, the model performance still outperforms DSN \cite{Lee2015Depply} using ground-truth labels as supervisory signals by 2.54\% (from 73.98\% to 76.52\%) and outperforms KD \cite{Lee2015Depply} only enabling the last student layer to learn class predictions by 0.97\% (from 75.55\% to 76.52\%).

We further explore the impact of the auxiliary classifier structure on the final performance. From the results of Table~\ref{table:aux_impact}, we can find that \textit{adding simple auxiliary classifiers} (a global average pooling layer and a fully-connected layer) to each shallow layer performs similarly to the baseline, which only allows the last layer to participate in the learning of teacher knowledge. This is because feature maps from shallow layers capture fine-grained details that are meaningless for the final prediction. It is thus necessary to use \textit{complex auxiliary classifiers} (adding multiple convolutional layers before the global average pooling layer) to capture coarse-grained features, which improves accuracy from 76.93\% to 78.05\%.

% \begin{table}
% \centering
% \caption{Ablation study with WRN-40-1 \& WRN-16-2.} 
% \resizebox{0.98\columnwidth}{!}{
% \begin{tabular}{ccccc}
% \hline\noalign{\smallskip}
% $\mathcal{L}_{Fea}$
% & & & \checkmark & \checkmark\\
% Adaptive weights & & \checkmark &  & \checkmark \\
% \hline\noalign{\smallskip}
% Test accuracy & 75.73$\pm$0.20 & 75.98$\pm$0.03 & 76.0$\pm$0.19 & \bf76.50$\pm$0.21\\
% \hline\noalign{\smallskip}
% \end{tabular}}
% \label{table:ablation}
% %\vspace{-0.3cm}
% \end{table}

\subsection{Weight Visualization}

\begin{figure}[t]
    \centering
    \centerline{\includegraphics[width=8.5cm]{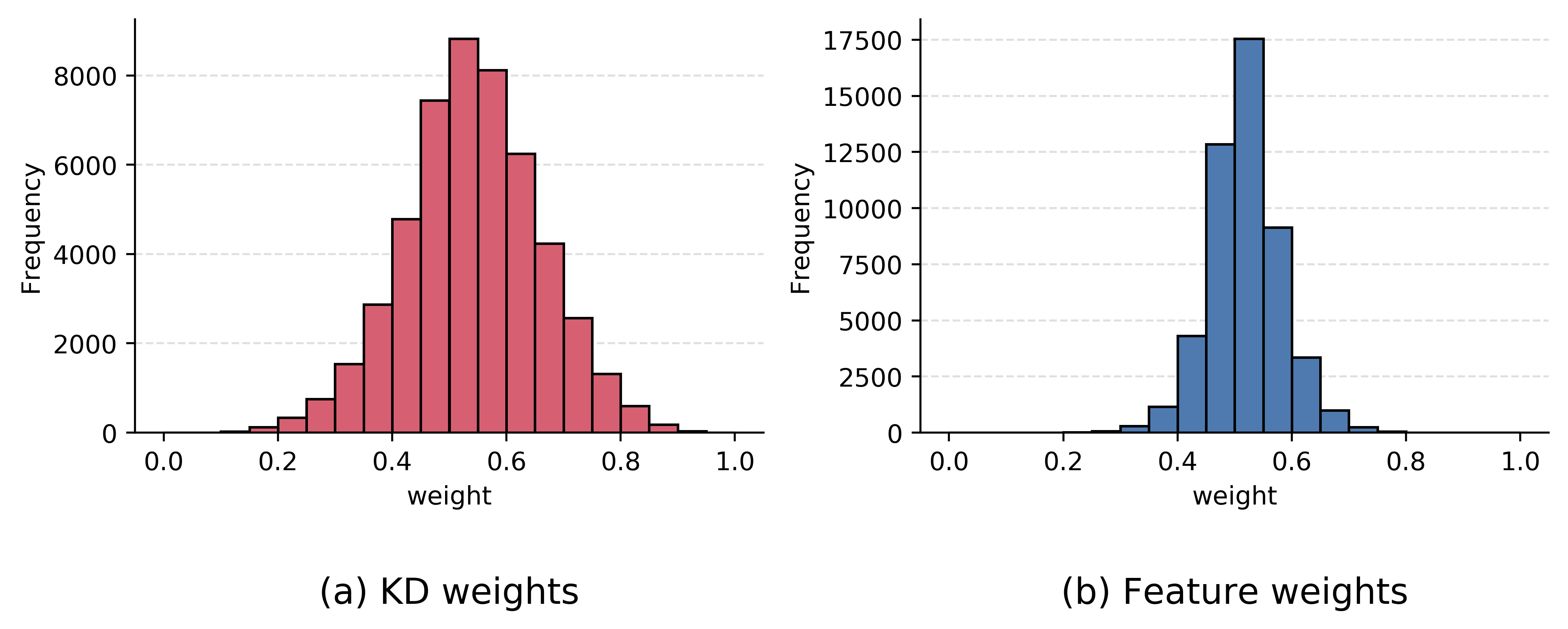}}
    \caption{The visualization of weight distribution on the first layer of ShuffleNetV2 supervised by ResNet32x4 on CIFAR-100. Weight represents loss-based weight, and frequency represents the number of samples.} 
    \label{fig:vis}
    %\vspace{-0.3cm}
\end{figure}

\begin{figure}[t]
    \centering
    \includegraphics[width=0.8\linewidth]{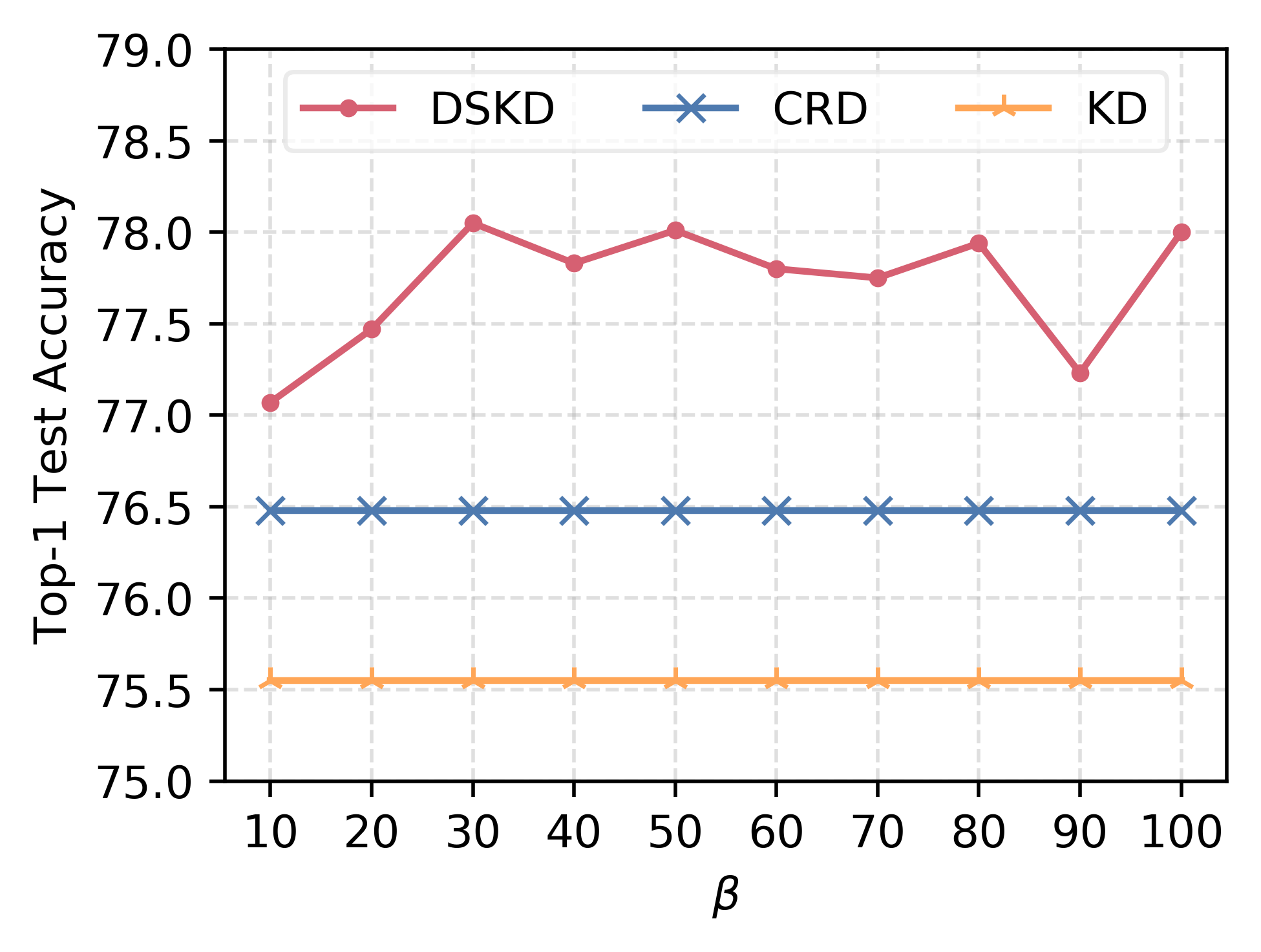}
    \caption{Sensitivity to $\beta$ on CIFAR-100 for ResNet32x4 \& ShuffleV2 } 
    \label{fig:hyper}
    %\vspace{-0.3cm}
\end{figure}

The adaptively learned weights for different samples on a certain layer are visualized in Fig.~\ref{fig:vis}. Since only two shallow layers used in ShuffleNetV2 and the weight sum of different layers equals to 1, we can easily infer the weight distribution on the second layer given Fig.~\ref{fig:vis}, i.e. flipping the figure with 180 degrees along the central axis. 

From the visualization results, we can observe that different weights are assigned to different samples, which hopefully help the student model training become better.

\subsection{Sensitivity Analysis}

We explore the impact of hyper-parameter $\beta$ on the performance of our DSKD method. 
As shown in Fig.~\ref{fig:hyper}, we plot Top-1 test accuracy on CIFAR-100 for ResNet32x4 \& ShuffleNetV2 with hyper-parameter $\beta$ ranging from 10 to 100 at equal interval of 10, and we compare with two KD methods: vanilla KD \cite{Hinton2015Distilling} and CRD \cite{Tian2020ContrastiveRD}. The blue and orange lines indicate the mean test accuracy of compared methods.

From this figure, we can find that our DSKD method achieves the best performance in all cases, which confirms the superiority of our proposed method. We also observe that our proposed method exhibits robust performance across various $\beta$ values, which means that DSKD can work well in a wide search space of hyper-parameter $\beta$.

\subsection{Computational Burden Analysis}

\begin{table}[t]
    \centering
     \caption{ Parameter count and computational cost of various models. (T), (S), and (+aux) refer to the teacher model, the student model, and the student model with auxiliary classifiers, respectively. }
    \begin{tabular}{ccc}
    \toprule
    & Params & Flops  \\
    \midrule
    WRN-40-2(T) & 2.25M & 330.42M \\
    WRN-16-2(S) & 0.70M & 102.10M \\
    WRN-16-2(+aux) & 0.76M & 104.97M \\
    \midrule
    Resnet32x4(T) & 7.43M & 1.08G \\
    Resnet8x4(S) & 1.23M & 178.57M \\
    Resnet8x4(+aux) & 1.40M & 189.05M \\
    \midrule
    WRN-40-2(T) & 2.25M & 330.42M \\
    MobileNetV2(S) & 0.81M & 7.37M \\
    MobileNetV2(+aux) & 0.89M & 7.65M \\
    \bottomrule
        
    \end{tabular}
   
    \label{table:burden}
\end{table}

Since we add a complex auxiliary classifier to each shallow layer of the student, a natural concern is whether this approach will cause a significant computational burden in the training period. But in fact, the additional convolutional layers in auxiliary classifiers are depthwise separable convolution to construct lightweight classifiers, which effectively reduces the additional computational burden. Table~\ref{table:burden} shows the changes in the parameters count and computational cost after adding auxiliary classifiers. Taking ``WRN-40-2 \& WRN-16-2" as an example, adding auxiliary classifiers just increases 8.5\% parameters (from 0.70M to 0.76M) and 2.8\% Flops (from 102.10M to 104.97M), which is much smaller than the teacher (params: 2.25M, Flops: 330.42M). So we consider this extra computational overhead to be totally acceptable. And these auxiliary classifiers are removed during inference period to obtain a final student model with low-memory and low-computation.

\section{Conclusion}
In this paper, we propose Deeply-Supervised Knowledge Distillation (DSKD) to make shallow layers of the student model participate in learning predictions from the teacher model, which further improves its final performance. We also develop a loss-based weight allocation strategy to balance the learning process of each shallow layer. Extensive experiments have demonstrated the effectiveness of our proposed method.

\bibliographystyle{IEEEtran}
\bibliography{IEEEexample}

\begin{thebibliography}{10}
\providecommand{\url}[1]{#1}
\csname url@samestyle\endcsname
\providecommand{\newblock}{\relax}
\providecommand{\bibinfo}[2]{#2}
\providecommand{\BIBentrySTDinterwordspacing}{\spaceskip=0pt\relax}
\providecommand{\BIBentryALTinterwordstretchfactor}{4}
\providecommand{\BIBentryALTinterwordspacing}{\spaceskip=\fontdimen2\font plus
\BIBentryALTinterwordstretchfactor\fontdimen3\font minus
  \fontdimen4\font\relax}
\providecommand{\BIBforeignlanguage}[2]{{%
\expandafter\ifx\csname l@#1\endcsname\relax
\typeout{** WARNING: IEEEtran.bst: No hyphenation pattern has been}%
\typeout{** loaded for the language `#1'. Using the pattern for}%
\typeout{** the default language instead.}%
\else
\language=\csname l@#1\endcsname
\fi
#2}}
\providecommand{\BIBdecl}{\relax}
\BIBdecl

\bibitem{Krizhevsky2012ImageNetCW}
A.~Krizhevsky, I.~Sutskever, and G.~E. Hinton, ``Imagenet classification with
  deep convolutional neural networks,'' in \emph{Advances in Neural Information
  Processing Systems}, 2012, pp. 1106--1114.

\bibitem{simonyan2015Very}
K.~Simonyan and A.~Zisserman, ``Very deep convolutional networks for
  large-scale image recognition,'' in \emph{International Conference on
  Learning Representations}, 2015.

\bibitem{He2016Deep}
K.~He, X.~Zhang, S.~Ren, and J.~Sun, ``Deep residual learning for image
  recognition,'' in \emph{Proceedings of the IEEE conference on computer vision
  and pattern recognition}, 2016, pp. 770--778.

\bibitem{Zagoruyko2016Wide}
S.~Zagoruyko and N.~Komodakis, ``Wide residual networks,'' in \emph{Proceedings
  of the British Machine Vision Conference}, 2016.

\bibitem{Hinton2015Distilling}
G.~E. Hinton, O.~Vinyals, and J.~Dean, ``Distilling the knowledge in a neural
  network,'' \emph{arXiv preprint arXiv:1503.02531}, 2015.

\bibitem{Gou2021KnowledgeDA}
J.~Gou, B.~Yu, S.~J. Maybank, and D.~Tao, ``Knowledge distillation: A survey,''
  \emph{International Journal of Computer Vision}, pp. 1789--1819, 2021.

\bibitem{Lee2015Depply}
C.-Y. Lee, S.~Xie, P.~Gallagher, Z.~Zhang, and Z.~Tu, ``Deeply-supervised
  nets,'' in \emph{Artificial intelligence and statistics}.\hskip 1em plus
  0.5em minus 0.4em\relax PMLR, 2015, pp. 562--570.

\bibitem{Sun2019Deeply}
D.~Sun, A.~Yao, A.~Zhou, and H.~Zhao, ``Deeply-supervised knowledge synergy,''
  in \emph{Proceedings of the IEEE/CVF Conference on Computer Vision and
  Pattern Recognition}, 2019, pp. 6997--7006.

\bibitem{Szegedy2015Going}
C.~Szegedy, W.~Liu, Y.~Jia, P.~Sermanet, S.~Reed, D.~Anguelov, D.~Erhan,
  V.~Vanhoucke, and A.~Rabinovich, ``Going deeper with convolutions,'' in
  \emph{Proceedings of the IEEE conference on computer vision and pattern
  recognition}, 2015, pp. 1--9.

\bibitem{gulccehre2016knowledge}
{\c{C}}.~G{\"u}l{\c{c}}ehre and Y.~Bengio, ``Knowledge matters: Importance of
  prior information for optimization,'' \emph{The Journal of Machine Learning
  Research}, 2016.

\bibitem{Romero2015FitNets}
R.~Adriana, B.~Nicolas, K.~S. Ebrahimi, C.~Antoine, G.~Carlo, and B.~Yoshua,
  ``Fitnets: Hints for thin deep nets,'' in \emph{International Conference on
  Learning Representations}, 2015.

\bibitem{Zagoruyko2017Paying}
N.~Komodakis and S.~Zagoruyko, ``Paying more attention to attention: improving
  the performance of convolutional neural networks via attention transfer,'' in
  \emph{International Conference on Learning Representations}, 2017.

\bibitem{Ahn2019Variational}
S.~Ahn, S.~X. Hu, A.~Damianou, N.~D. Lawrence, and Z.~Dai, ``Variational
  information distillation for knowledge transfer,'' in \emph{Proceedings of
  the IEEE/CVF Conference on Computer Vision and Pattern Recognition}, 2019,
  pp. 9163--9171.

\bibitem{Chen2021Cross}
D.~Chen, J.-P. Mei, Y.~Zhang, C.~Wang, Z.~Wang, Y.~Feng, and C.~Chen,
  ``Cross-layer distillation with semantic calibration,'' in \emph{Proceedings
  of the AAAI Conference on Artificial Intelligence}, 2021, pp. 7028--7036.

\bibitem{Ji2021Show}
M.~Ji, B.~Heo, and S.~Park, ``Show, attend and distill: Knowledge distillation
  via attention-based feature matching,'' in \emph{Proceedings of the AAAI
  Conference on Artificial Intelligence}, 2021, pp. 7945--7952.

\bibitem{chen2020online}
D.~Chen, J.-P. Mei, C.~Wang, Y.~Feng, and C.~Chen, ``Online knowledge
  distillation with diverse peers,'' in \emph{Proceedings of the AAAI
  Conference on Artificial Intelligence}, 2020, pp. 3430--3437.

\bibitem{chen2022knowledge}
D.~Chen, J.-P. Mei, H.~Zhang, C.~Wang, Y.~Feng, and C.~Chen, ``Knowledge
  distillation with the reused teacher classifier,'' in \emph{Proceedings of
  the IEEE/CVF Conference on Computer Vision and Pattern Recognition}, 2022,
  pp. 11\,933--11\,942.

\bibitem{Tian2020ContrastiveRD}
Y.~Tian, D.~Krishnan, and P.~Isola, ``Contrastive representation
  distillation,'' in \emph{International Conference on Learning
  Representations}, 2020.

\bibitem{Liu2021ExploringIC}
L.~Liu, Q.~Huang, S.~Lin, H.~Xie, B.~Wang, X.~Chang, and X.~Liang, ``Exploring
  inter-channel correlation for diversity-preserved knowledge distillation,''
  in \emph{Proceedings of the IEEE/CVF International Conference on Computer
  Vision}, 2021, pp. 8271--8280.

\bibitem{lin2022knowledge}
S.~Lin, H.~Xie, B.~Wang, K.~Yu, X.~Chang, X.~Liang, and G.~Wang, ``Knowledge
  distillation via the target-aware transformer,'' in \emph{Proceedings of the
  IEEE/CVF Conference on Computer Vision and Pattern Recognition}, 2022, pp.
  10\,915--10\,924.

\bibitem{Tung2019SimilarityPreservingKD}
F.~Tung and G.~Mori, ``Similarity-preserving knowledge distillation,'' in
  \emph{Proceedings of the IEEE/CVF International Conference on Computer
  Vision}, 2019, pp. 1365--1374.

\bibitem{Yim2017AGF}
J.~Yim, D.~Joo, J.~Bae, and J.~Kim, ``A gift from knowledge distillation: Fast
  optimization, network minimization and transfer learning,'' in
  \emph{Proceedings of the IEEE conference on computer vision and pattern
  recognition}, 2017, pp. 4133--4141.

\bibitem{Lin2017FocalLF}
T.-Y. Lin, P.~Goyal, R.~Girshick, K.~He, and P.~Doll{\'a}r, ``Focal loss for
  dense object detection,'' in \emph{Proceedings of the IEEE international
  conference on computer vision}, 2017, pp. 2980--2988.

\bibitem{Zhang2018ExFuseEF}
Z.~Zhang, X.~Zhang, C.~Peng, X.~Xue, and J.~Sun, ``Exfuse: Enhancing feature
  fusion for semantic segmentation,'' in \emph{Proceedings of the European
  conference on computer vision}, 2018, pp. 269--284.

\bibitem{Newell2016StackedHN}
A.~Newell, K.~Yang, and J.~Deng, ``Stacked hourglass networks for human pose
  estimation,'' in \emph{Proceedings of the European conference on computer
  vision}, 2016.

\bibitem{Zhang2019BYOT}
L.~Zhang, J.~Song, A.~Gao, J.~Chen, C.~Bao, and K.~Ma, ``Be your own teacher:
  Improve the performance of convolutional neural networks via self
  distillation,'' in \emph{Proceedings of the IEEE/CVF International Conference
  on Computer Vision}, 2019, pp. 3713--3722.

\bibitem{Li2020OnlineKD}
Z.~Li, Y.~Huang, D.~Chen, T.~Luo, N.~Cai, and Z.~Pan, ``Online knowledge
  distillation via multi-branch diversity enhancement,'' in \emph{Proceedings
  of the Asian Conference on Computer Vision}, 2020.

\bibitem{Chen2019DropAO}
Y.~Chen, H.~Fan, B.~Xu, Z.~Yan, Y.~Kalantidis, M.~Rohrbach, S.~Yan, and
  J.~Feng, ``Drop an octave: Reducing spatial redundancy in convolutional
  neural networks with octave convolution,'' in \emph{Proceedings of the
  IEEE/CVF International Conference on Computer Vision}, 2019, pp. 3435--3444.

\bibitem{Huang2018Multi}
G.~Huang, D.~Chen, T.~Li, F.~Wu, L.~Van Der~Maaten, and K.~Q. Weinberger,
  ``Multi-scale dense networks for resource efficient image classification,''
  in \emph{International Conference on Learning Representations}, 2017.

\bibitem{Zeiler2014VisualizingAU}
M.~D. Zeiler and R.~Fergus, ``Visualizing and understanding convolutional
  networks,'' in \emph{European conference on computer vision}, 2014, pp.
  818--833.

\bibitem{Chollet2017XceptionDL}
F.~Chollet, ``Xception: Deep learning with depthwise separable convolutions,''
  in \emph{Proceedings of the IEEE conference on computer vision and pattern
  recognition}, 2017, pp. 1251--1258.

\bibitem{Huang2022KnowledgeDF}
T.~Huang, S.~You, F.~Wang, C.~Qian, and C.~Xu, ``Knowledge distillation from a
  stronger teacher,'' in \emph{Conference on Neural Information Processing
  Systems}, 2022.

\bibitem{Skorski2021Revisiting}
M.~Skorski, A.~Temperoni, and M.~Theobald, ``Revisiting weight initialization
  of deep neural networks,'' in \emph{Asian Conference on Machine Learning},
  2021, pp. 1192--1207.

\bibitem{krizhevsky2009learning}
A.~Krizhevsky, G.~Hinton \emph{et~al.}, ``Learning multiple layers of features
  from tiny images,'' \emph{Technical Report}, 2009.

\bibitem{Le2015TinyIV}
Y.~Le and X.~Yang, ``Tiny imagenet visual recognition challenge,''
  \emph{Technical Report}, 2015.

\bibitem{Sandler2018MobileNetV2}
M.~Sandler, A.~Howard, M.~Zhu, A.~Zhmoginov, and L.-C. Chen, ``Mobilenetv2:
  Inverted residuals and linear bottlenecks,'' in \emph{Proceedings of the IEEE
  conference on computer vision and pattern recognition}, 2018, pp. 4510--4520.

\bibitem{zhang2018shufflenet}
X.~Zhang, X.~Zhou, M.~Lin, and J.~Sun, ``Shufflenet: An extremely efficient
  convolutional neural network for mobile devices,'' in \emph{Proceedings of
  the IEEE/CVF Conference on Computer Vision and Pattern Recognition}, 2018.

\bibitem{Ma2018shufflenv2}
N.~Ma, X.~Zhang, H.-T. Zheng, and J.~Sun, ``Shufflenet v2: Practical guidelines
  for efficient cnn architecture design,'' in \emph{Proceedings of the European
  conference on computer vision}, 2018, pp. 116--131.

\bibitem{yao2020knowledge}
A.~Yao and D.~Sun, ``Knowledge transfer via dense cross-layer
  mutual-distillation,'' in \emph{European Conference on Computer Vision},
  2020, pp. 294--311.

\bibitem{Paszke2019PyTorchAI}
A.~Paszke, S.~Gross, F.~Massa, A.~Lerer, J.~Bradbury, G.~Chanan, T.~Killeen,
  Z.~Lin, N.~Gimelshein, L.~Antiga \emph{et~al.}, ``Pytorch: An imperative
  style, high-performance deep learning library,'' in \emph{Advances in neural
  information processing systems}, 2019.

\bibitem{van2008visualizing}
L.~Van~der Maaten and G.~Hinton, ``Visualizing data using t-sne.''
  \emph{Journal of machine learning research}, 2008.

\end{thebibliography}

\end{document}